\documentclass[10pt,twocolumn,letterpaper]{article}
\usepackage{authblk}
\usepackage[pagenumbers]{cvpr} % To force page numbers, e.g. for an arXiv version

\usepackage{graphicx}
\usepackage{amsmath}
\usepackage{amssymb}
\usepackage{booktabs}

\usepackage[pagebackref,breaklinks,colorlinks]{hyperref}

\usepackage[capitalize]{cleveref}
\crefname{section}{Sec.}{Secs.}
\Crefname{section}{Section}{Sections}
\Crefname{table}{Table}{Tables}
\crefname{table}{Tab.}{Tabs.}

\def\cvprPaperID{764} % *** Enter the CVPR Paper ID here
\def\confName{WACV}
\def\confYear{2023}

\begin{document}

%%%%%%%%% TITLE - PLEASE UPDATE
\title{3DMM-RF: Convolutional Radiance Fields for 3D Face Modeling}

\newcommand{\ourname}{3DMM-RF}

\author[1,2]{Stathis Galanakis}

% For a paper whose authors are all at the same institution,
% omit the following lines up until the closing ``}''.
% Additional authors and addresses can be added with ``\and'',
% just like the second author.
% To save space, use either the email address or home page, not both
\author[2]{Baris Gecer}
\author[1,2]{Alexandros Lattas}
\author[1,2]{Stefanos Zafeiriou}

\affil[1]{Imperial College of London}
\affil[2]{Huawei}
% \affil[3]{\tt\small s.galanakis21@imperial.ac.uk}}
% \affil[4]{\tt\small baris.gecer@huawei.com}
% \affil[5]{\tt\small alexandros.lattas17@imperial.ac.uk}
% \affil[6]{\tt\small s.zafeiriou@imperial.ac.uk}

\maketitle

%%%%%%%%% ABSTRACT
\begin{abstract}
Facial 3D Morphable Models are a main computer vision subject with countless applications
and have been highly optimized in the last two decades.
The tremendous improvements of deep generative networks have created various possibilities for improving such models and have attracted wide interest.
Moreover, the recent advances in neural radiance fields, are revolutionising novel-view synthesis of known scenes.
In this work, we present a facial 3D Morphable Model,
which exploits both of the above, 
and can accurately model a subject's identity, pose and expression
and render it in arbitrary illumination.
This is achieved by utilizing a powerful deep style-based generator to overcome two main weaknesses of neural radiance fields,
their rigidity and rendering speed.
We introduce a style-based generative network that synthesizes in one pass
all and only the required rendering samples of a neural radiance field.
We create a vast labelled synthetic dataset of facial renders,
and train the network on these data, so that it can accurately model and generalize on facial identity, pose and appearance.
Finally, we show that this model can accurately be fit to ``in-the-wild'' facial images
of arbitrary pose and illumination, extract the facial characteristics,
and be used to re-render the face in controllable conditions.
\end{abstract}

%%%%%%%%% BODY TEXT
\section{Introduction}
Photorealistic 3D face modeling and reconstruction from a single image is a widely researched field in computer vision due to its numerous applications, such as avatar creation, virtual makeup and speech-driven face animation. Since the introduction of 3D Morphable Models (3DMM) in 1999 by the seminal work of Blanz and Vetter~\cite{Blanz1999AMM} to model faces by statistical linear models, the majority of the research stroll around linear representation of faces. Many follow-up works have studied the integration of 3DMMs with Deep Neural Networks~\cite{tewari2019fml,tran2019learning}, Mesh Convolutions~\cite{moschoglou20193dfacegan}, and Generative Adversarial Networks (GANs)~\cite{gecer2019ganfit,gecer2020fastganfit,gecer2020synthesizing} in order to improve its representational strength in high-frequency details and photorealism. In this study, we explore the potential of recently emerged non-linear representation approach called Neural Radiance Fields (NeRFs)~\cite{mildenhall2020nerf} to evolve 3DMMs into its strong neural volumetric representation.

NeRFs~\cite{mildenhall2020nerf} have recently demonstrated an immense progress in novel-view synthesis ~\cite{mildenhall2020nerf}, relighting ~\cite{9578828, 9710856}, and reconstruction ~\cite{Chan2021}.
They consist of fully connected neural networks that learn to implicitly represent a scene and its appearance parameters.
Such networks can be optimized by using only a few dozens camera views of the scene and can be queried to generate novel views of that scene.
Despite their photorealistic renderings and high 3D consistency, most initial NeRF-based methods focus
on modeling and overfitting on a single subject, at a single pose.

On the contrary, methods like the 3D Morphable Model(3DMM)~\cite{Blanz1999AMM},
create a statistical model of human faces by hundreds or thousands of 3D scans, and can be used as a prior to reconstructing a 3D face from a single image, with controllable identity, pose and appearance.
Despite their flexibility 
and the tremendous improvements in their ability to generate photorealistic faces~\cite{gecer2020synthesizing},
they are difficult to re-create authentic facial renderings
and their rendering relies on expensive skin shading models~\cite{lattas2021avatarme++}.

Currently, there has been tremendous progress in high-resolution 2D image generation, using generative adversarial networks~\cite{Karras2019stylegan2, Karras2020ada}.
Such methods have been shown to successfully generate photorealistic human faces~\cite{Karras2019stylegan2, Karras2020ada}.
Extensions have been proposed, which combined with a 3DMM, 
can even generate facial images with controllable facial, camera and environment attributes~\cite{tewari2020stylerig}.
However, the domain of such methods lies in the 2D space
and attempting to freely interpolate camera or illumination parameters creates unwanted artifacts.

% Furhtermore, NeRFs generally predict the voxel information given the coordinates by employing a simple Multi Layer Perceptron (MLP) which is computationally inefficient and disregard spatial correlation of the representation. 
In this work, we create an implicit 3D Morphable Model,
by leveraging a powerful style-based deep generator network,
with a flexible radiance field representation and volume rendering module.
Since NeRFs heavily overfit to a single scene, they are not constrained by the capacity of the neural component. On the other hand, generic models, such GANs and 3DMM, compress information fed into the model from hundreds of thousands of data samples. In order to build a generic NeRF representation, we propose an architecture that can predict the whole volumetric representation in a single inference. The architecture utilizes deconvolutional layers as in a traditional generator network, thus provides spatial consistency, computational and memory efficiency so that it can model a large-scale dataset.
In order to unlock the full potential of our approach, we generate a synthetic dataset including 10,000 photorealisticly rendered 3D faces with identity, expression, pose and illumination variations. Therefore, the proposed volumetric representation model can disentangle attributes such as identity, expression, illumination and pose, exposing an advantage over single scene/object~\cite{mildenhall2020nerf} and large-scale unsupervised~\cite{chanefficienta} approaches. 
In summary, our main contributions are:
\begin{itemize}
\itemsep-0.5em 
    \item We present \ourname, a controllable parametric face model inspired by Neural Radiance Fields.
    \item We show that the proposed \ourname{}  model can be used for 3D reconstruction of ``in-the-wild`` face images.
    \item Our convolutional generator architecture with depth-based sampling strategy
    can generate the samples required for rendering a volumetric radiance field,
    in a single pass.
    \item We introduce a large-scale synthetic dataset that disentangles identity, expression, camera and illumination.
\end{itemize}

%-------------------------------------------------------------------------
\section{Related Work}

\subsection{3D Face Modeling}
Ever since the original 3D Morphable Model introduced by~\cite{Blanz1999AMM}, there have been many studies in that direction such as extension to facial expressions~\cite{cao2013facewarehouse,yang2011expression,li2017learning,breidt2011robust,amberg2008expression,li2010example,thies2015real,bouaziz2013online}, large-scale dataset releases~\cite{paysan20093d,booth20163d,smith2020morphable,dai20173d}, reconstruction by deep regression networks~\cite{tran2017regressing,tewari2017mofa,r.2020monocular,gecer2020fastganfit}, and reconstruction by analysis-by-synthesis with more advanced features~\cite{booth20173d,gecer2019ganfit,gecer2020synthesizing}.
Due to its linear nature, the original 3DMMs  under-represent the high-frequency information and often result in overly-smoothed geometry and texture models. In terms of preserving photorealism and high-frequency signals, non-linear generative models have been shown to be very successful in 2D image synthesis~\cite{goodfellow2014generative,kingma2014autoencoding,karras2017progressive,vahdat2020nvae}, thus, non-linear 3D face modeling has been widely studied in the context of deep generative networks~\cite{tewari2019fml,tran2019learninga,tewari2017self}, GANs~\cite{gecer2019ganfit,gecer2020synthesizing,lattas2020avatarme}, and VAEs~\cite{bagautdinov2018modeling,lombardi2018deep,wei2019vr,ranjan2018generating}.
We refer the reader to ~\cite{zollhofer2018state,3dfacereconstructionsurvey}
for a more detailed presentation of the 3D face modeling approaches.

%-------------------------------------------------------------------------

\subsection{Neural Radiance Fields}
Original NeRF paper~\cite{mildenhall2020nerf} showed promising results for becoming an improvement to the conventional scene representations e.g. point-clouds, meshes. This and its extensions~\cite{ 9578784, kaizhang2020} can implicitly represent scenes by using a differentiable neural rendering algorithm, without the need of 3D supervision. One of the drawbacks of NeRF approaches is the computationally heavy rendering process and its long training time. Many approaches have been introduced focusing on reducing training and inference time~\cite{9710808, 9711398, 9710021, 9710464}.
Other approaches focus on 3D shape reconstruction~\cite{yariv2021volume, 9709919, wang2021neus}, human bodies registration~\cite{peng2021neural, weng2022humannerf, peng2021animatable}, dealing with non-static scenes~\cite{9578753,9711476}, scene editing~\cite{9710143, zhang2021stnerf} and scene relighting~\cite{9578828, 9710856}.
NeRF-based networks' applications have been extended on representing human faces based on a video~\cite{9578714, 9711476,  guo2021adnerf, Wang_2021_CVPR, park2021hypernerf} or a single image~\cite{Gao-portraitnerf,zhuang2021mofanerf, hong2021headnerf}.
The recent work i3DMM~\cite{yenamandra2020i3dmm} is the first to represent a 3DMM using an implicit function, using signed distance functions (SDF).
Our method can be categorized as the latter type of networks and its application is quite similar to~\cite{zhuang2021mofanerf} and~\cite{hong2021headnerf}. Our main distinction is the fact that both of them perform a low-dimension rendering and then improve the image quality through upsampling layers~\cite{hong2021headnerf} or by a Refine network~\cite{zhuang2021mofanerf}, whereas our method doesn't require any optimization after the rendering step.

%-------------------------------------------------------------------------

\subsection{Deep Generative Models}
The impressive photorealistic results of the GAN paper~\cite{NIPS2014_5ca3e9b1} resulted in its widespread application~\cite{DBLP:journals/corr/abs-1710-10196,3dfacereconstructionsurvey, DBLP:journals/corr/abs-1812-04948,
Karras2019stylegan2, Karras2020ada}.
Currently, there is a large effort of combining NeRF approaches with GANs starting with pi-GAN~\cite{9577547} and
GRAF~\cite{DBLP:journals/corr/abs-2007-02442}.
Most recent approaches like GIRAFFE~\cite{Niemeyer2020GIRAFFE}, StyleNerf~\cite{gu2022stylenerf}, CIPS-3D~\cite{zhou2021CIPS3D} and EG-3D~\cite{Chan2021} have shown remarkable image quality results. 
GIRAFFE~\cite{Niemeyer2020GIRAFFE} and StyleNerf~\cite{gu2022stylenerf} are two-stage networks, having a Multi-Layer Perceptron (MLP) at low
resolution and then upsampling, CIPS-3D~\cite{zhou2021cips} synthesizes each pixel independently whereas EG-3D~\cite{Chan2021} introduces a 3D aware generative model base on the tri-plane representation.
A distinction of our method is that these approaches use an MLP which predicts feature vectors while ours renders RGB$\alpha$ output directly. Also, some of these approaches require a two-step sampling procedure, while \ourname{} uses a depth-based sampling strategy, which needs only a single pass.
%------------------------------------------------------------------------

\begin{figure*}
\begin{center}
\includegraphics[width=1.0\linewidth]
                  {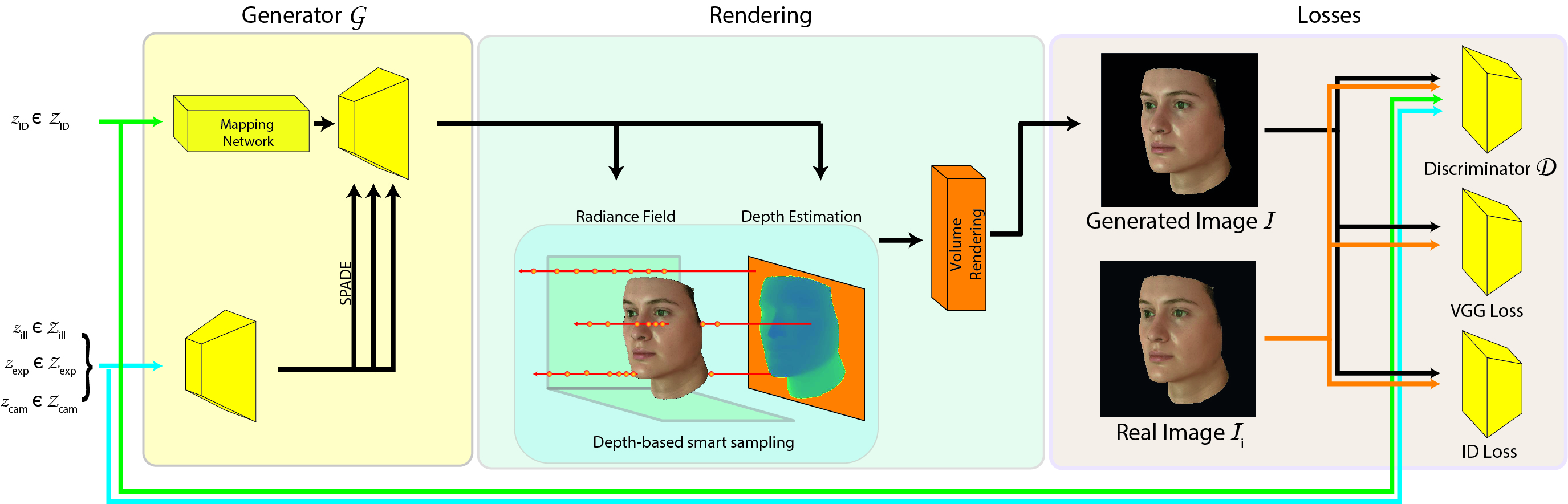}
\end{center}
  \caption{
  Overview of \ourname{},
  a generative network which creates the sampling space of a neural radiance field depicting a human face, in one pass, based on the depths acquired from the depth prediction and then renders an image directly using the volume rendering algorithm~\cite{mildenhall2020nerf}.
  The Generator $\mathcal{G}$ is conditioned on a subject's identity vector $\textbf{z}_{ID}$, as well as the expression parameters $\textbf{z}_{exp}$,
  the scene illumination parameters $\textbf{z}_{ill}$ and the camera parameters $\textbf{z}_{cam}$.
  $\textbf{z}_{ID}$ passes through a mapping network, while
  the rest are transformed by SPADE layers before being given as input to $\mathcal{G}$.
  Based on a ray-face intersection depth estimation, the generator directly generates the samples needed for volume rendering which speeds up rendering.
  Finally, the network is trained on a synthetic dataset,
  using adversarial, perceptual (VGG) and identity losses.
  }
\label{fig:overview}
\end{figure*}

\section{3D Face Model by Neural Radiance Fields}

In this work, we describe an implicit parametric facial model,
with disentangled identity, pose and appearance.
The NeRF representation is unfit for such a task, as
a) it represents a rigid scene,
b) optimizing a scene with a large number of identities,
poses and appearance requires an intractable optimization. 
In this manner, we introduce \ourname{}, 
a model that can represent and render a controllable non-rigid facial scene,
using a style-based generator~\cite{Karras2019stylegan2},
that generates an instance of an implicit neural radiance field.
Moreover, \ourname{} learns to approximate the area of dense samples for each view, so that a rendering can be achieved with a single query of the network.

Considering a 3D space as a neural rendering space, 
\ourname{} is a neural morphable model $\mathcal{S}$ that renders a facial image $\mathbf{I} \in \mathbb{R}^{512\times 512 \times 3}$ as follows:
\begin{equation}
    \mathbf{I} = \mathcal{S}\left(\textbf{z}_{ID}, \textbf{z}_{exp}, \textbf{z}_{cam}, \textbf{z}_{ill}\right)
\end{equation}
where $\textbf{z}_{ID} \in \mathbb{R}^{512}$ describes an identity latent code, $\textbf{z}_{exp} \in \mathbb{R}^{20}$ the expression 3DMM blendshapes, $\textbf{z}_{cam} \in \mathbb{R}^3$ the camera position, whereas $\textbf{z}_{ill} \in \mathbb{R}^8$ the illumination parameters.
The \ourname{} model $\mathcal{S}$ consists of style-based generator $\mathcal{G}$ that generates a volumetric radiance field, a volume rendering module, and a discriminator $\mathcal{D}$.
An overview of the method is shown in Fig.~\ref{fig:overview},
the network architecture is presented in Sec.~\ref{section:network},
the training using the synthetic dataset in Sec.~\ref{section:parameters},
and the fitting process in Sec.~\ref{section:fitting}.

%------------------------------------------------------------------------

\subsection{The Architecture}
\label{section:network}

\subsubsection{Scene Representation}
Consider a neural radiance field (NeRF)~\cite{mildenhall2020nerf} with an implicit function 
$F_{\Theta}(x,y,z,\theta,\phi)\longrightarrow (r,g,b,\sigma)$,
that maps a world coordinate $(x,y,z)$ and a viewing direction $(\theta,\phi)$,
to a colour $(r,g,b)$ with density $\sigma$.
A rigid scene can be implicitly learned by the radiance field,
by fitting $F_{\theta}$ on different views of the scene.
To render a novel view, $F_{\theta}$ is queried using hierarchical sampling, for a number of points on each ray intersecting the camera, 
and volume rendering~\cite{mildenhall2020nerf} is used to produce the final colour.

Contrary to the typical NeRF approach~\cite{mildenhall2020nerf}
which implicitly represents a scene with a trained MLP,
we train a convolutional generator network $\mathcal{G}$ (Sec.~\ref{sec:generator})
that concurrently generates all necessary $K$ samples for each ray that passes through each pixel of the rendered image $I$, in a single pass.
Each sample contains colour $r,g,b$ and density $\sigma$ values, like the sample vectors in NeRF~\cite{mildenhall2020nerf}.

\subsubsection{Depth-based Smart Volume Rendering}
NeRF rendering is plagued by its computational load since, typically per pixel,
firstly a uniform sampling step is required to find the densest area,
and another to extract a dense sampling area for volume rendering.
For our approach, a given ray \emph{r} will terminate as soon as it intersects with the facial surface. This means that samples, which are a bit far from the surface, will not contribute to the final RGB values. Based on that idea, we significantly speed-up rendering, by predicting the ray-face intersection depth $D_{r}$, for the ray $r$ and sample only around this area. The predicted depth $D_{r}$ is modelled as a Gaussian distribution with mean $D_{\mu_{r}}$ and standard variation $D_{std_{r}}$.
Overall, for a ray $r$ containing $K$ samples, 
our model generates $N$ channels, where $N = 4K + 2$, which the first \emph{4K} channels represent each sample across the ray $r$,
and two additional channels for the depth prediction. Throughout the layers of the generator, sample values (\emph{4K}) evolve together with depth estimation, 
meaning that they are highly correlated thanks to the ground truth provided by the synthetic data. 
Thus, generated sampled values are aligned with the estimated depth values in the final radiance field.

% Following the aforementioned strategy, we acquire the necessary depth values for all the rays. 
% Moreover, our model generates all the required samples along the ray, 
% at a single pass.
Therefore, by generating only all the required samples along the ray that are close to the facial surface, we can directly employ the volume rendering, bypassing
the multiple MLP queries required by NeRF and its importance sampling strategy \cite{mildenhall2020nerf}.
For the single ray \emph{r}, we separate the depth prediction channels
and reshape the rest into $K \times 4$ samples along the ray.
Each sample $k \in 1\dots{}K$,
includes the predicted $\mathbf{c_k} = (r_k,g_k,b_k)$ colour values
and their corresponding density $\sigma_k$.
Using the predicted depth mean $D_{\mu_{r}}$ and standard variation $D_{std_{r}}$ to sample the depth values,
the final colour $\hat{\textbf{C}}_r$ is measured using standard volume rendering \cite{mildenhall2020nerf}:
\begin{equation}
    \begin{split}
        \hat{\textbf{C}}_r & = \sum_{k=1}^{K} w_k \textbf{c}_k, \\
        w_k & = T_k \left(1 - exp \left( -\sigma_k \left(t_{k+1} - t_{k} \right) \right) \right), \\ 
        T_n & = exp \left( - \sum_{k=1}^{K-1} \sigma_k \left(t_{k+1} - t_{k} \right) \right), \\
        t_i &\sim \mathcal{N}(D_{\mu_r}, D_{std_r}) \text{~and~} t_{k+1} \geq t_{k}
    \end{split}
\end{equation}
$w_k$ is each sample's contribution and 
$t_k$ is its depth.

\vspace{-0.3cm}
\subsubsection{Convolutional Radiance Field Generator}

\label{sec:generator}
\begin{figure*}[t]
\centering
    \vspace{-1cm}
    \includegraphics[width=1.0\linewidth]
                      {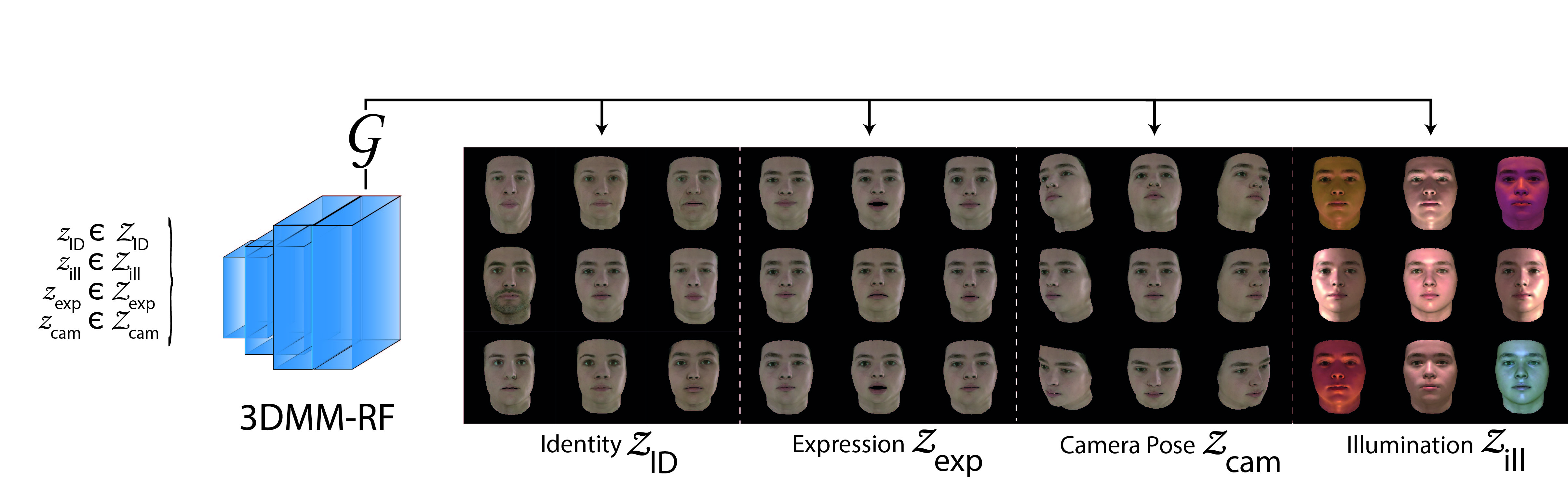}
    \label{fig:capabilities}
    \vspace{-0.5cm}
    \caption{
        Once trained, \ourname{} has learned disentangled latent representations for facial attributes and scene parameters,
        and can easily be queried to render facial images.
        From left to right, we show samples of
        a) subject identity $\textbf{z}_{ID} \in Z_{ID}$,
        b) subject expression $\textbf{z}_{exp} \in Z_{exp}$,
        c) camera pose $\textbf{z}_{cam} \in Z_{cam}$,
        and d) scene illumination $\textbf{z}_{ill} \in Z_{ill}$ .
    }
\end{figure*}
% \vspace{-1cm}

We extend the above single-ray prediction approach,
to the prediction of all the rays required to render an image $\mathbf{I} \in \mathbb{R}^{512 \times 512 \times 3}$.
In this manner, we introduce a generator network $\mathcal{G}$ that generates all the required samples of neural radiance field in a single pass,
as it is shown in Fig.~\ref{fig:overview}.
The capacity of the deep convolutional generator enables it to generate the radiance field of arbitrary scenes, conditioned on certain attributes.
The generator consists of a fully-convolutional mapping network~\cite{Karras2019stylegan2},
which translates a subject's identity vector $\textbf{z}_{ID}$ to an extended latent space identity vector $w_{ID}$,
and a synthesis network that generates all the ray samples of interest.

In addition to the extended latent space identity vector $w_{ID}$,
we condition the generator with properties relating to the reconstructed scene, namely 
the blendshape expression vector $\textbf{z}_{exp}$,
the camera position $\textbf{z}_{cam}$
and the scene illumination parameters $\textbf{z}_{ill}$,
which include the light source direction, the diffuse, specular and  ambient intensity.
% For the camera position and the light source direction, 
We also apply positional encoding~\cite{mildenhall2020nerf} to  the camera position and the light source direction vectors.
% since we empirically notice that it helps the network interpolate between identities more smoothly. 
We observed that the generator $G$ performs better when it receives the scene parameters 
as a 2-dimensional vector instead of 1-dimensional.
This is achieved by firstly passing the parameters $\textbf{z}_{exp}, \textbf{z}_{cam}, \textbf{z}_{ill}$
through two modulated convolutional layers~\cite{Karras2019stylegan2}
which are then fed to the network through SPADE layers~\cite{park2019SPADE}.

Following the typical adversarial training strategy~\cite{goodfellow2014generative, Karras2019stylegan2}, a discriminator $\mathcal{D}$ is needed to achieve photorealistic results.
We follow the architecture of the StyleGAN2~\cite{Karras2019stylegan2} discriminator 
using the conditional strategy introduced by
~\cite{Karras2020ada}. 
Moreover, the discriminator is conditioned by both identity $\textbf{z}_{ID}$ and the rest parameters $\textbf{z}_{exp}, \textbf{z}_{cam}, \textbf{z}_{ill}$, which are fed to it as a conditional label.

%------------------------------------------------------------------------

\subsection{Training by Synthetic Face Dataset} \label{section:parameters}
\begin{figure}[h!]
    \centering
    \includegraphics[width=\linewidth]{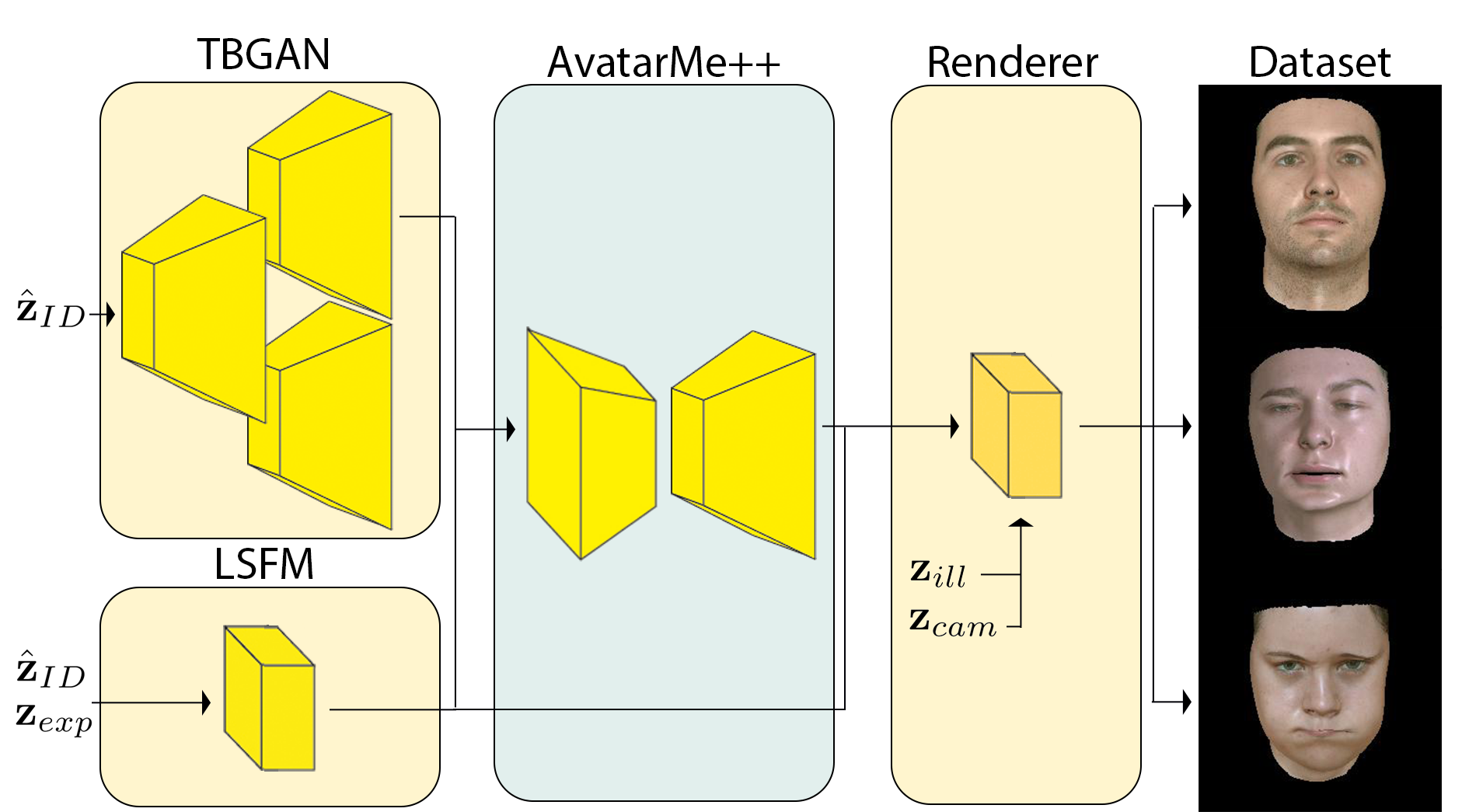}
    \caption{
    Overview of our dataset generation pipeline:
    For each training image, 
    its identity vector $\mathbf{\hat{z}}_{ID}$ and expression vector $\mathbf{z}_{exp}$
    are used to generate diverse facial textures and shapes with TBGAN \cite{gecer2020synthesizing}
    and LSFM \cite{booth2018large}.
    Then AvatarMe++ \cite{lattas2021avatarme++} is used to acquire reflectance textures,
    which are photorealistically rendered with PyTorch3D \cite{ravi2020pytorch3d},
    given then illumination $\mathbf{z}_{ill}$ and camera $\mathbf{z}_{cam}$ vectors.
    }
    % \vspace{-1cm}
    \label{fig:dataset}
\end{figure}

% intro
Given the lack of accurately labelled facial images 
with paired depth maps and labelled variations in illumination, expression and pose, 
we train the proposed architecture using a synthetically generated dataset.
This vast training dataset consists of arbitrary photorealistically rendered facial images, which are based on a 3D Moprhable Model. They are also paired with metadata concerning their characteristics and rendering,
information difficult to acquire for ``in-the-wild'' images.
Despite using synthetic data, our generator's finetuning step (Sec.~\ref{section:fitting}) accurately captures real facial images.

% Dataset
We generate a synthetic dataset using a morphable model (LSFM \cite{booth2018large}), a facial texture generator (TBGAN \cite{gecer2020synthesizing}) and a facial material network (AvatarMe++ \cite{lattas2021avatarme++}).
Specifically, we draw 10,000 facial shape samples from LSFM,
and another 10,000 facial texture samples from TBGAN based on an identity vector. These samples are purposefully uncorrelated to increase subject diversity.
However, the textures lack photorealistically rendering material properties,
and thus we pass them through an image-to-image translation network (AvatarMe++),
which translates each facial texture to spatially varying
(a) diffuse albedo, (b) specular albedo, (c) diffuse normals and (d) specular normals.
As shown in \cite{lattas2021avatarme++}, these can be used for photorealistically rendering facial images under arbitrary environment conditions using shaders such as the Blinn-Phong \cite{phong},
which we implement in PyTorch3D~\cite{ravi2020pytorch3d, lattas2021avatarme++}.
Moreover, we sample various real expression blendshapes $\mathbf{Z}_{exp}$ from 4DFAB~\cite{cao2013facewarehouse}, which correspond to our 3DMM.
Finally, we define a space of the environment illumination $\mathbf{Z}_{ill}$ with plausible RGB values and random direction for $n_l$ light sources, and plausible RGB values for ambient illumination,
and a space of the frontal-hemisphere rendering camera parameters $\mathbf{Z}_{cam}$ (up to $30^\circ$ on each axis).
The models we use \cite{gecer2020synthesizing, booth2018large, lattas2021avatarme++}
are trained on diverse datasets (see each paper for details),
and their trained models or training data are in the public domain.

% params
During training, we use the above to generate vast amounts of training data.
The complete dataset rendering function is defined as 
$\mathcal{R}(\mathbf{\hat{z}}_{ID}, \mathbf{z}_{exp}, \mathbf{z}_{ill}, \mathbf{z}_{cam}) \xrightarrow{} \mathbb{R}^{w,h,4}$,
where $\hat{\mathbf{z}}_{ID}$ is an identity vector,
$\mathbf{z}_{exp}$ is an expression vector,
$\mathbf{z}_{ill}$ is an illumination vector,
$\mathbf{z}_{cam}$ is a camera vector and
$w,h$ is the image shape.
The output's 3 channel are the RGB rendering facial image,
and the last channel is the rasterized camera-space depth,
using SoftRasterizer \cite{liu2019soft}.
For each training iteration \emph{i}, we generate each labelled image and its depth,
$\mathcal{R}(\mathbf{\hat{z}}_{{ID}_i}, \mathbf{z}_{{exp}_i}, \mathbf{z}_{{ill}_i}, \mathbf{z}_{{cam}_i}) = \left\{\mathbf{I_i}, \mathbf{D_i}\right\}$,
by generating a random identity vector $\mathbf{\hat{z}}_{{ID}_i}$,
and drawing arbitrary expression blendshapes $\mathbf{z}_{{exp}_i} \in \mathbf{Z}_{exp}$, camera params $\mathbf{z}_{{cam}_i} \in \mathbf{Z}_{cam}$
and illumination params $\mathbf{z}_{{ill}_i} \in \mathbf{Z}_{ill}$
from their sets.
Finally, to increase compatibility with ``in-the-will'' images,
we use a state-of-the-art face recognition network~\cite{deng2019arcface} on the rendered image $\mathbf{I_i}$,
to acquire the latent identity vector $\mathbf{z}_{{ID}_i}$ 
which we give to our model instead of $\mathbf{\hat{z}}_{{ID}_i}$,
to facilitate the use of an identity loss.

% \paragraph{Lighting}

% \paragraph{Camera Position}

%------------------------------------------------------------------------
%% \subsection{Full-control of     %

% We may write something about the results -> full control, expression lighting %

%------------------------------------------------------------------------

\subsection{Fitting for 3D Face Reconstruction} \label{section:fitting}

We can utilize our \ourname{} model to 3D reconstruct any single ``in-the-wild`` image.
Given any face image, our goal is to get the optimal latent vectors for identity $\textbf{z}_{ID}$, expression $\textbf{z}_{exp}$, camera position $\textbf{z}_{cam}$, and illumination $\textbf{z}_{ill}$ that can
reconstruct the target face. Firstly, we align the target face by using a face detector network~\cite{guo2021sample} and also acquire its identity latent vector $\tilde{\textbf{z}}_{ID}$ by a face recognition network~\cite{deng2018arcface}.
Then by using a facial landmark detection network~\cite{bulat2017far}, we extract the 2D face landmarks and get a face mask based on these landmarks.  We initialize the input identity latent vector $\textbf{z}_{ID}$ as the identity latent vector $\tilde{\textbf{z}}_{ID}$,
and set the expression parameters $\textbf{z}_{exp}$ to a zero vector, camera position $\textbf{z}_{cam}$ to a frontal view and illumination $\textbf{z}_{ill}$ to an average illumination setting.
During the optimization, we first freeze our network and optimize the 
$\mathbf{z}$ parameters where $\mathbf{z} = \left\{ \textbf{z}_{ID},\textbf{z}_{exp},
\textbf{z}_{cam}, \textbf{z}_{ill} \right\}$, by comparing the final rendered image and the target image by a combination of loss functions as follows:
\begin{equation}
  L_{fitting} = L_{pht} + L_{vgg} + L_{ID} + L_{landmarks}
\end{equation}
where $L_{pht}$ is the MSE loss (Photometry) between the rendered and the final image, 
$L_{vgg}$ is the perceptual loss as introduced in~\cite{zhang2018perceptual}, $L_{ID}$ is the loss between the identity feature maps based on~\cite{deng2018arcface} and $L_{landmarks}$ is the
facial landmarks loss. We define $L_{landmarks}$ as the $L2-$ distance between the activation maps of both images which are fed into the facial landmark network~\cite{bulat2017far}.
To further optimize the reconstruction of the face, we finetune the generator network $\mathcal{G}$ parameters in addition to the input parameters $\mathbf{z}$ by the same loss function with 200-fold smaller learning rate. This approach helps to recover identity more precisely and helps to bridge the domain gap between the synthetic training data and ``in-the-wild`` images.
%------------------------------------------------------------------------
\section{Experiments}
%------------------------------------------------------------------------

\begin{figure*}[t]
\begin{center}
  \centering
    \begin{subfigure}[b]{0.12\textwidth}
         \centering
         \includegraphics[height=6.8cm]{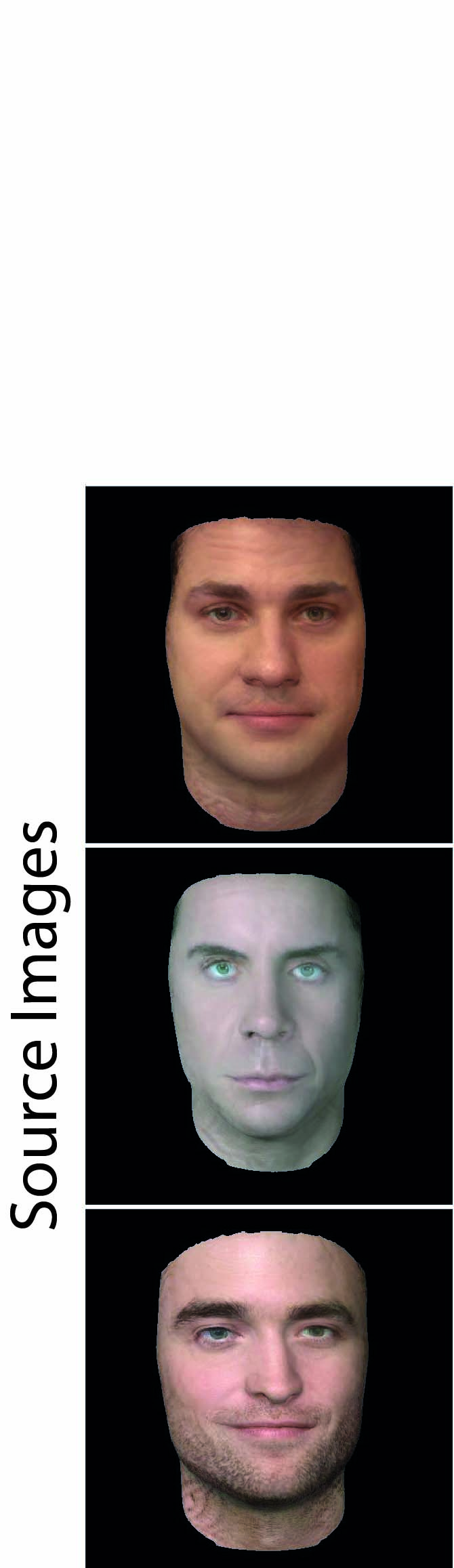}
         \caption{Source}
         \label{fig:changing_input_img}
     \end{subfigure}
    \begin{subfigure}[b]{0.28\textwidth}
     \centering
     \includegraphics[height=6.8cm]{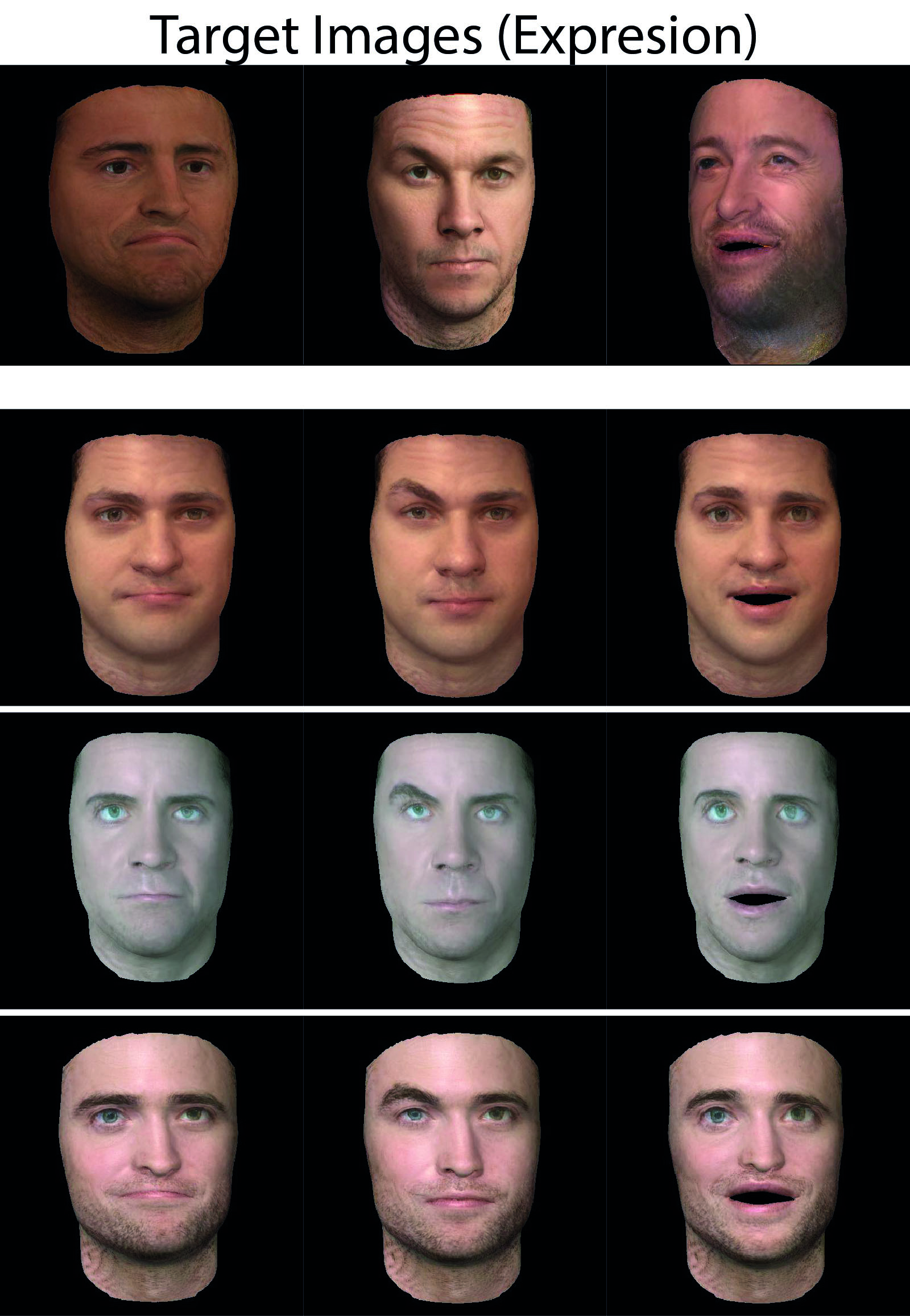}
         \caption{
            Expression transfer ($\textbf{z}_{exp}$)
         }
         \label{fig:changing_exp}
     \end{subfigure}
    \begin{subfigure}[b]{0.28\textwidth}
     \centering
     \includegraphics[height=6.8cm]{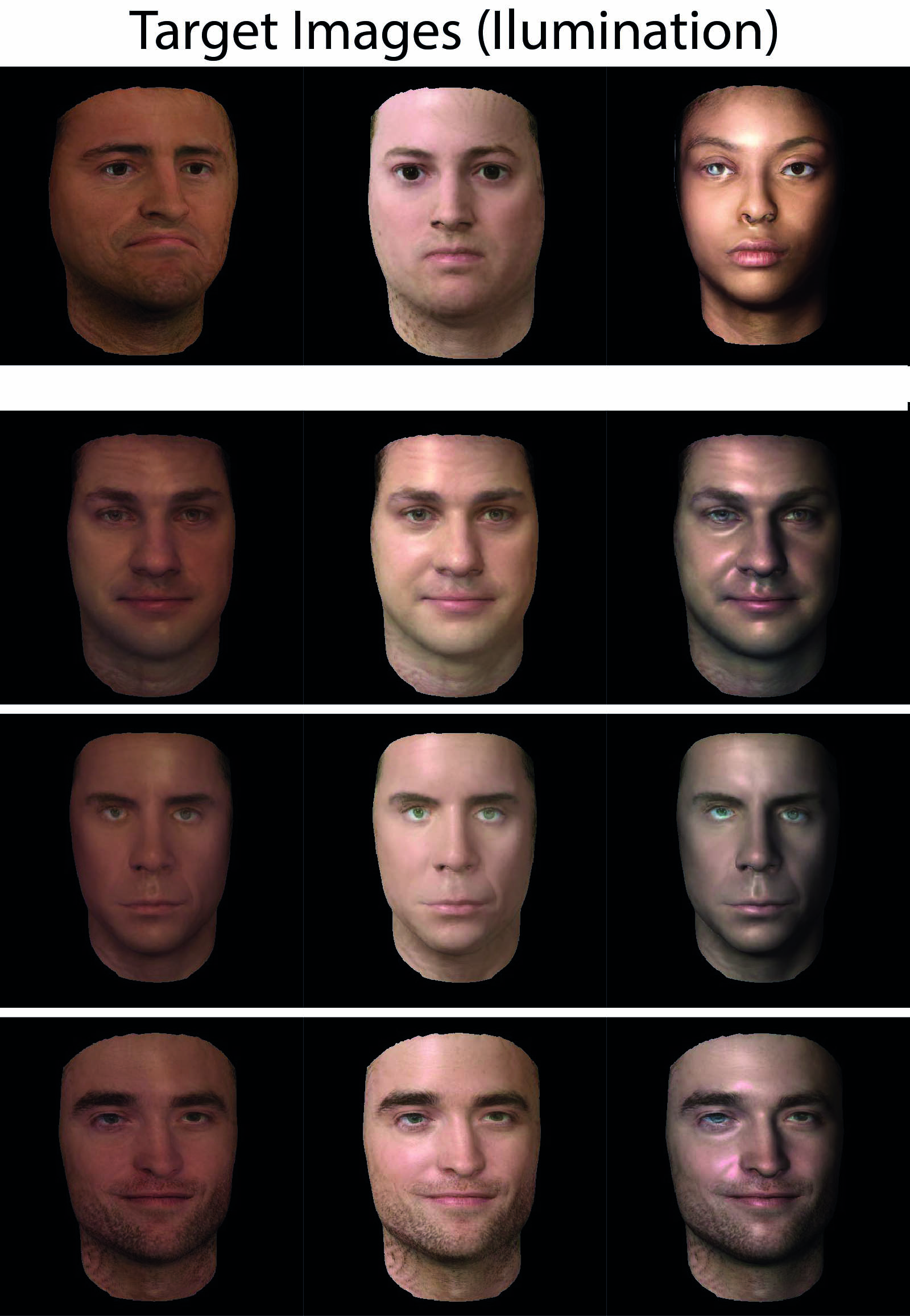}
         \caption{
            Illumination transfer ($\textbf{z}_{ill}$)
         }
         \label{fig:changing_ill}
     \end{subfigure}
    \begin{subfigure}[b]{0.28\textwidth}
     \centering
     \includegraphics[height=6.8cm]{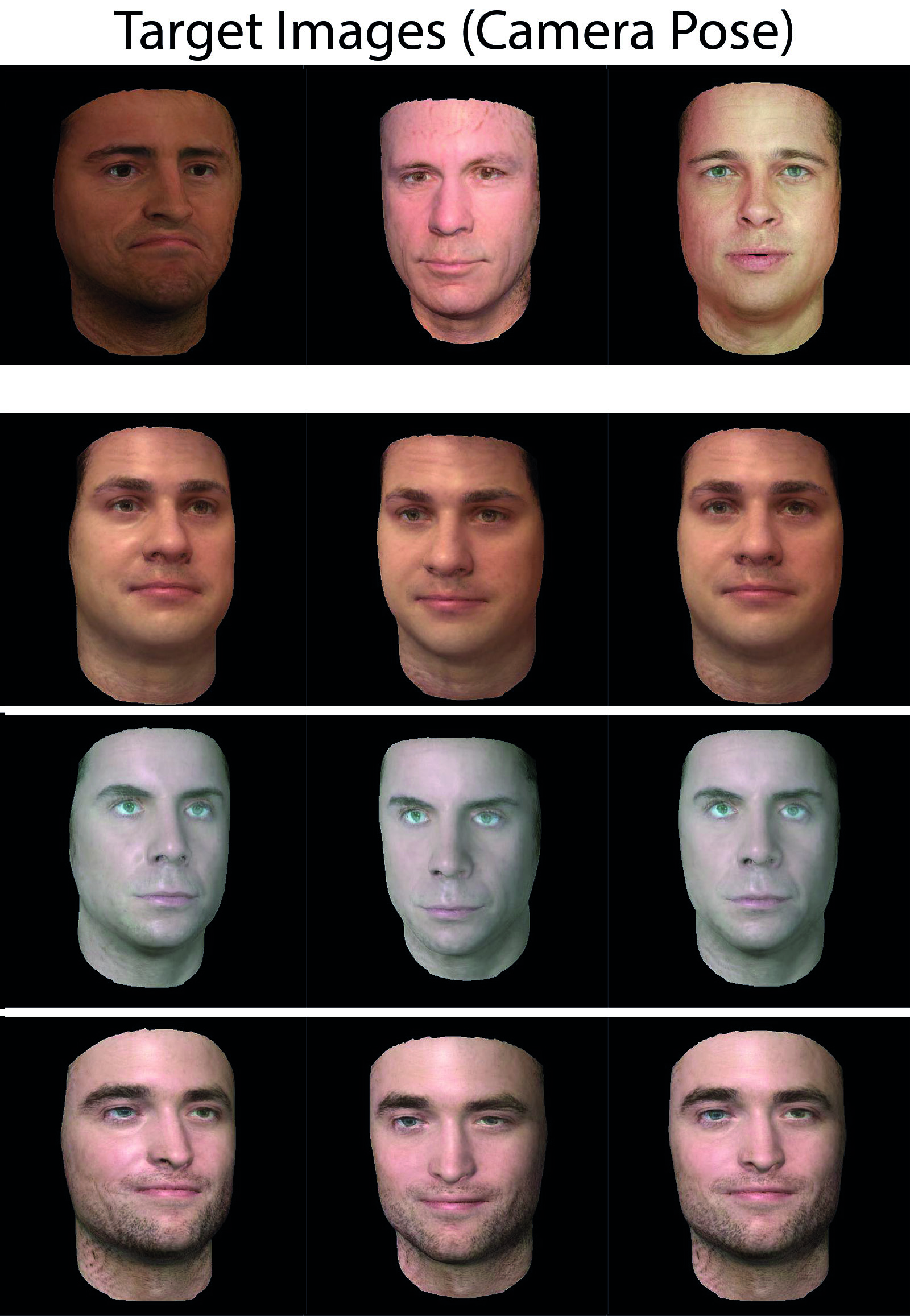}
         \caption{
            Camera pose transfer ($\textbf{z}_{cam}$)
         }
         \label{fig:changing_cam}
     \end{subfigure}
\caption{
    We showcase the ability of our network to disentangle facial and scene properties, 
    by reconstructing the source images on the left,
    and replacing some of their attributes, with those acquired from target images.
}
\label{fig:multi_scene}
\end{center}
\end{figure*}

\subsection{Disentanglement control}

One of the goals of the training is the network to be able to efficiently disentangle its latent space. This means that
each dimension of the resulting subspace affects a different variation factor.
% In other words, for a  combination of latent vectors $\textbf{z}_id, z_expr, z_cam, z_il$, while modifying one of the aforementioned vectors e.g. $\textbf{z}_ill$, StyleFaN preserves the identity, the expression and the camera pose of the rendered face.
As shown in Fig.~\ref{fig:multi_scene}, while modifying one of the input parameters, our synthesis network is capable of preserving the 
rest unchanged. Given the input images of Fig. \ref{fig:changing_input_img}, the renderings in Fig. 
\ref{fig:changing_exp} demonstrate that \ourname{} achieves expression transfer from the desired target images to the input ones, while it preserves the same  identities, lighting conditions and camera poses in each case. 
%------------------------------------------------------------------------
\subsection{Comparison with other models}

\def \var {0.1205}
\begin{figure*}[t]
\begin{center}
  \centering
    \begin{subfigure}[t]{\var\textwidth}
         \centering
         \includegraphics[width=1.0\textwidth]{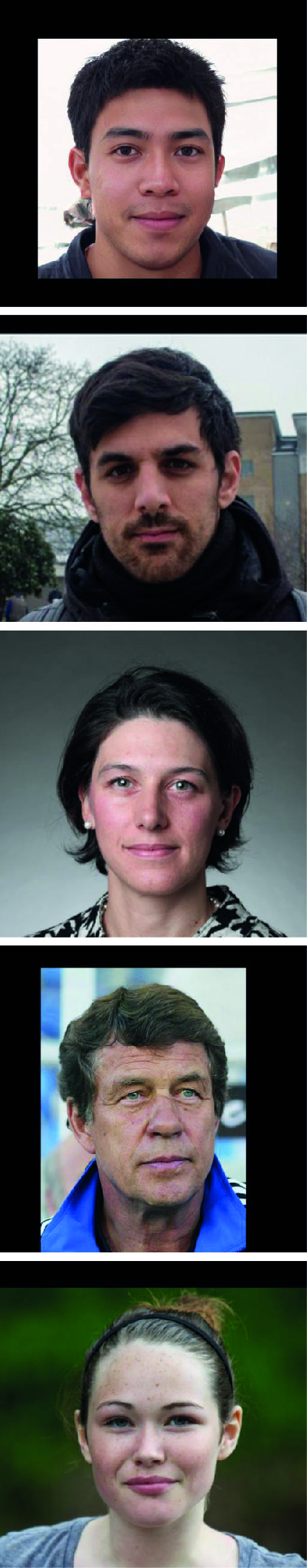}
         \caption{Input \\ Image}
         \label{fig:multi_com_input}
     \end{subfigure}
    \begin{subfigure}[t]{\var\textwidth}
     \centering
     \includegraphics[width=1.0\textwidth]{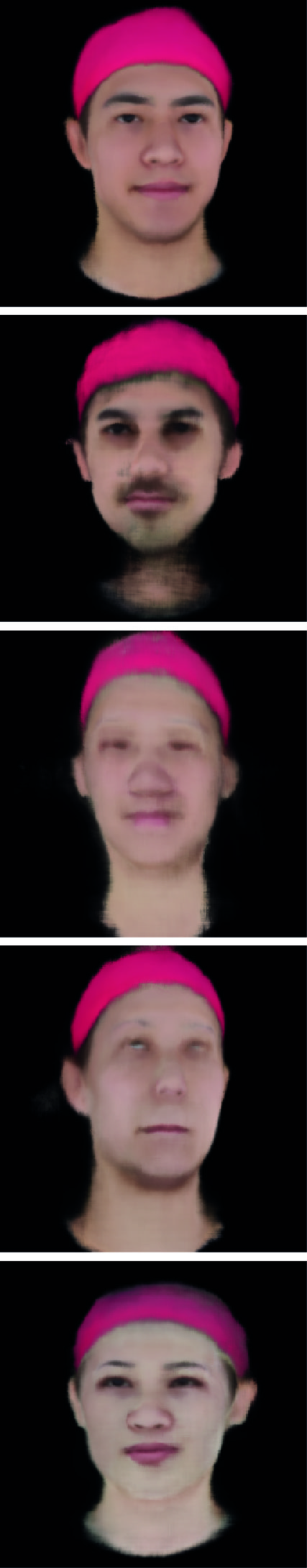}
         \caption{ \centering MoFaNeRF~\cite{zhuang2021mofanerf}}
         \label{fig:multi_com_mofa}
     \end{subfigure}
    \begin{subfigure}[t]{\var\textwidth}
     \centering
     \includegraphics[width=1.0\textwidth]{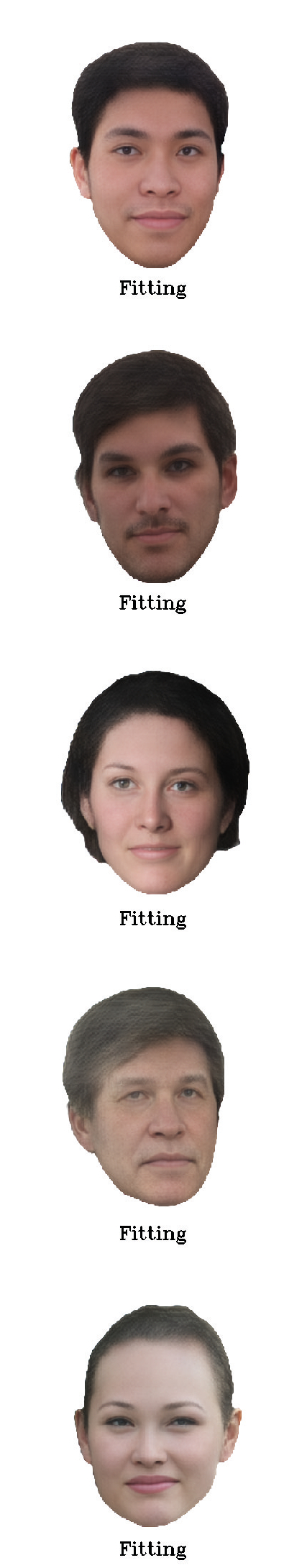}
         \caption{\centering HeadNeRF~\cite{hong2021headnerf}}
         \label{fig:multi_com_headnerf}
     \end{subfigure}
    \begin{subfigure}[t]{\var\textwidth}
     \centering
     \includegraphics[width=1.0\textwidth]{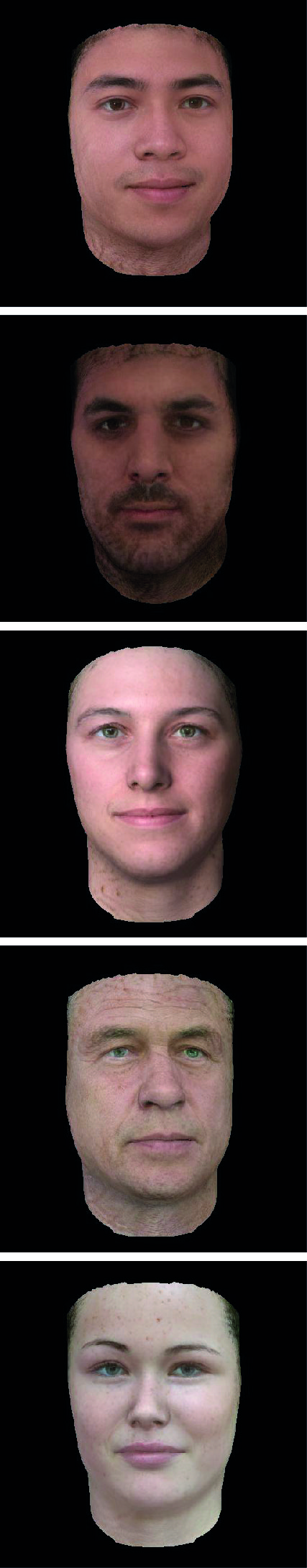}
         \caption{\centering Ours (\ourname{})}
         \label{fig:multi_com_ours}
     \end{subfigure}
    \begin{subfigure}[t]{\var\textwidth}
     \centering
     \includegraphics[width=1.0\textwidth]{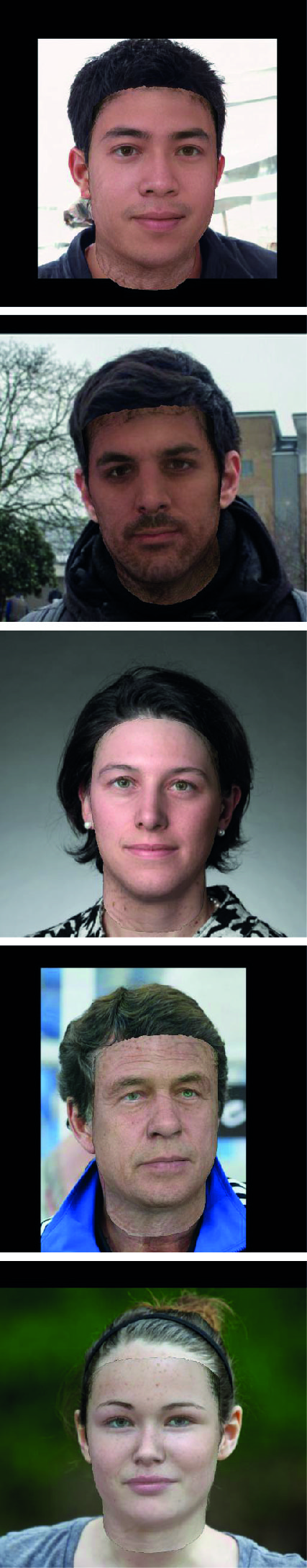}
         \caption{ \centering Ours overlaid}
         \label{x}
     \end{subfigure}
    \begin{subfigure}[t]{\var\textwidth}
     \centering
     \includegraphics[width=1.0\textwidth]{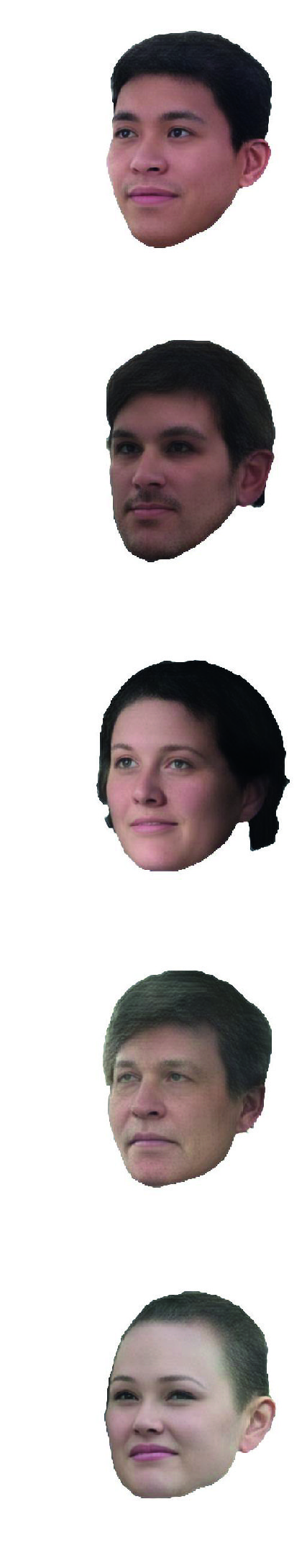}
         \caption{ \centering HeadNeRF Novel views}
         \label{fig:multi_com_novel_headnerf}
     \end{subfigure}
    \begin{subfigure}[t]{\var\textwidth}
     \centering
     \includegraphics[width=1.0\textwidth]{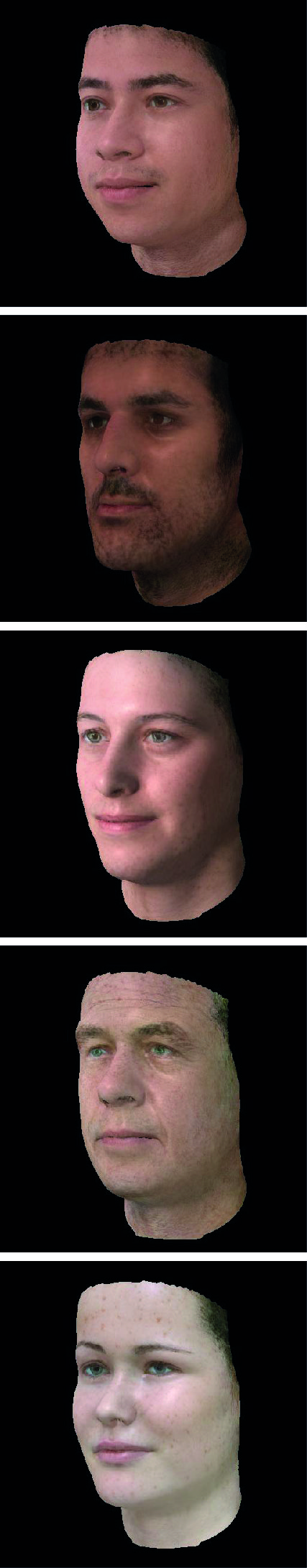}
         \caption{Ours \\ Novel views}
         \label{fig:multi_com_novel_ours}
     \end{subfigure}
    \begin{subfigure}[t]{\var\textwidth}
     \centering
     \includegraphics[width=1.0\textwidth]{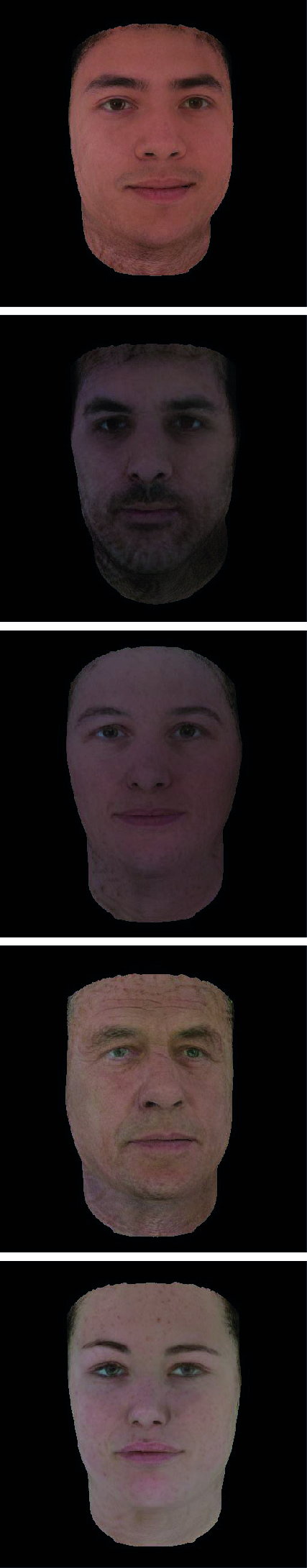}
         \caption{\centering Diffuse albedo}
         \label{fig:multi_com_albedo}
     \end{subfigure}
\caption{This figure contains a qualitative comparison between our method (\ourname{}), MoFaNeRF~\cite{zhuang2021mofanerf} and HeadNeRF~\cite{hong2021headnerf}. From left to right, the columns include the input image, MoFaNeRF's fitted prediction, HeadNeRF's prediction, \ourname{}'s prediction, a novel view rendered by HeadNeRF and \ourname{} and ours under diffuse albedo. }
\label{fig:comparison}
\end{center}
\vspace{-0.4cm}
\end{figure*}

\begin{figure*}
\centering
    \includegraphics[width=1.0\textwidth]{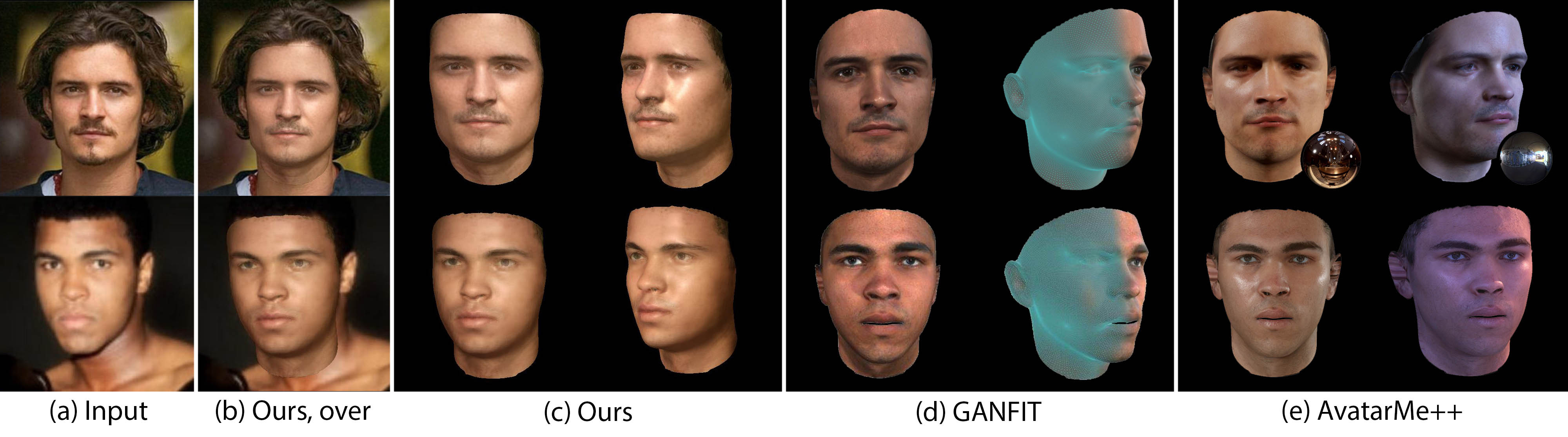}
    \vspace{-0.5cm}
    \caption{
        Comparison between our results and explicit 3DMM fitting methods, GANFIT \cite{Gecer_2019_CVPR} and AvatarMe++ \cite{lattas2021avatarme++}.
        We showcase that our reconstruction and rendering are on-par with these state-of-the-art methods, despite being implicit.
    }
    \label{fig:comp_ganfit_avatarme}
    % \vspace{-0.5cm}
\end{figure*}

% We compare our method with other similar state-of-the-art methods for evaluating its effectiveness qualitatively and quantitatively.

\subsubsection{Qualitative comparison}
We compare the fitting ability of our method with  MoFaNeRF~\cite{zhuang2021mofanerf} and HeadNeRF~\cite{hong2021headnerf}, both NeRF-based face models. MoFaNeRF is a parametric model which maps face images into a vector space including identity, expression and appearance features. HeadNeRF is another NeRF-based parametric head model achieving photorealistic results.
Fig. \ref{fig:comparison} shows the fitting results for the same input images, which are presented in Fig.~\ref{fig:multi_com_input}. Even though MoFaNeRF appears to perform well(Fig.~\ref{fig:multi_com_mofa}), in some cases it completely misses to recreate a human face. Moreover, although HeadNeRF achieves to reproduce more photorealistic human faces than MoFaNeRF does(Fig.~\ref{fig:multi_com_headnerf}), it doesn't always capture the right identity. It is clearly shown that, except rows a\&e, the final rendered identity is not the same as the one appearing in the input image. As shown in Fig.~\ref{fig:multi_com_ours}, our approach achieves both high-quality output while it preserves the right identity.
Even for rows a\&e where HeadNerf manages to reconstruct the right identity, our approach produces finer details. 

Moreover,
we compare our implicit 3DMM modeling and rendering,
with two state-of-the-art explicit 3DMM fitting methods,
GANFIT \cite{Gecer_2019_CVPR} and AvatarMe++ \cite{lattas2021avatarme++}.
Fig.~\ref{fig:comp_ganfit_avatarme} shows our results on two subjects,
from a central and side pose, compared to the above methods.
Even though above methods \cite{Gecer_2019_CVPR, lattas2021avatarme++} use explicit rendering based on a mesh,
we showcase that our reconstruction is on-par with such methods,
and our neural facial rotation accurately maintains the facial and environment characteristics.

\vspace{-0.4cm}

\subsubsection{Quantitative comparison}

% \vspace{-0.25cm}

\begin{table}
\begin{center}
\begin{tabular}{|l|c|c|c|}
\hline
Method & L1 $\downarrow$ & LPIPS $\downarrow$  & SSIM $\uparrow$ \\
\hline\hline
MoFaNeRF~\cite{zhuang2021mofanerf} & 0.37 & 0.098 & 0.9155 \\
\hline
HeadNeRF~\cite{hong2021headnerf} & 0.241 & 0.05 & 0.9506 \\
\hline
\ourname~(Ours) & \textbf{0.216} & \textbf{0.035} & \textbf{0.9563}\\
\hline
\end{tabular}
\end{center}
\caption{Quantitative comparison between \ourname (Ours), MoFaNeRF~\cite{zhuang2021mofanerf} and HeadNeRF~\cite{hong2021headnerf} using image comparison metric }
\label{table:comparison_metrics}
\vspace{-0.25cm}
\end{table}

We measure the ability of our network of preserving the identity features by reproducing the steps followed in~\cite{gecer2019ganfit} and ~\cite{genova2018unsupervised}.
Firstly, we reconstruct every image of the Labeled Faces in the Wild Dataset (LFW)~\cite{Huang2007a} using our method and then we feed them both to a face recognition network~\cite{BMVC2015_41}. Then, we compare the cosine similarity distributions of the activations in the network's embedding layer between pairs of 1) the original and the reconstructed-rendered image in Fig.~\ref{fig:id_comparison} 2) images containing the same/different identities in Fig.~\ref{fig:id_pairs}.
We show that the reconstructed images have more than 0.75 cosine similarity to the original images in average, and same/different pairs of LFW dataset are still distinguishable after our reconstruction. Both results show that identities are well-preserved. Our performance closely matches with the other state-of-the-art methods that are specializing in identity preservation such as~\cite{genova2018unsupervised,gecer2019ganfit,gecer2020fastganfit,tran2017disentangled}.

Additionally, we randomly select 550 images from the CelebAMask-HQ~\cite{CelebAMask-HQ} following the method introduced by HeadNeRF~\cite{hong2021headnerf}. We do fit our model on those images and compare the fitted images with the state-of-the-art models MoFaNeRF~\cite{zhuang2021mofanerf} and HeadNeRF~\cite{hong2021headnerf}. In contrast with the approach proposed by the authors of HeadNeRF, we do not train any of the networks using the rest images contained in this dataset. 
% The reason is that we want to evaluate the ability of our model to fit in completely unseen images, without any bias introduced by a dataset. 
As comparison metrics, we use  $L1$-distance, SSIM~\cite{ssim} and LPIPS~\cite{zhang2018perceptual}. The results are shown in  Tab.~\ref{table:comparison_metrics} which clearly shows that our method outperforms.

\def \var {0.23}
\begin{figure}[t]
\begin{center}
% \fbox{\rule{0pt}{2in} \rule{0.9\linewidth}{0pt}}
\begin{subfigure}[b]{\var\textwidth}
     \centering
     \includegraphics[width=1.0\textwidth]{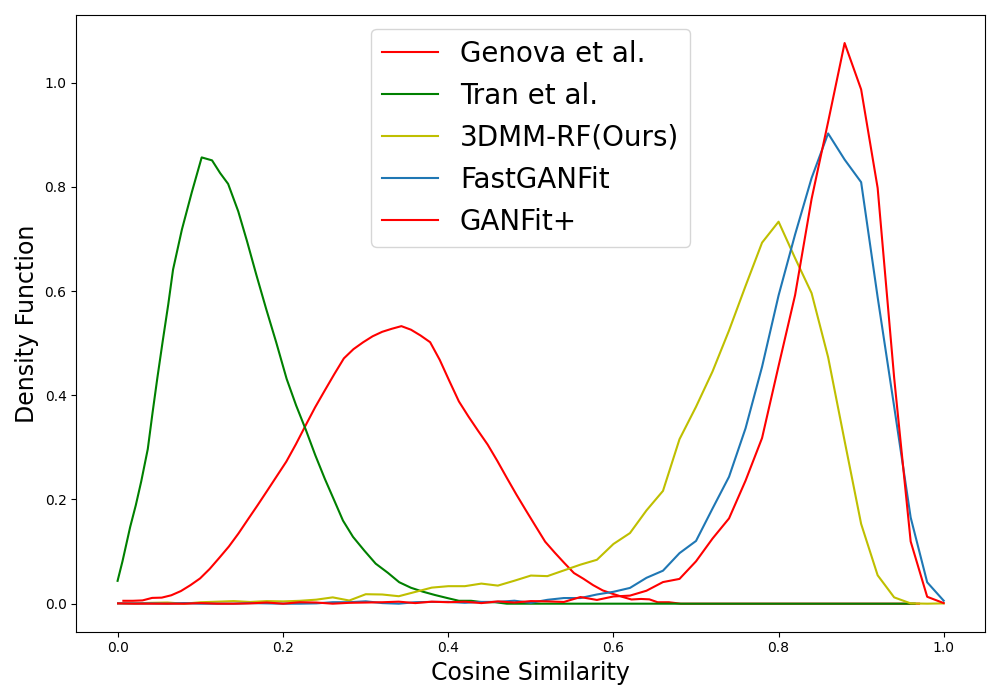}
         \caption{\centering  Identity preservation of fitted images}
         \label{fig:id_comparison}
     \end{subfigure}
     \begin{subfigure}[b]{\var\textwidth}
     \centering
     \includegraphics[width=1.0\textwidth]{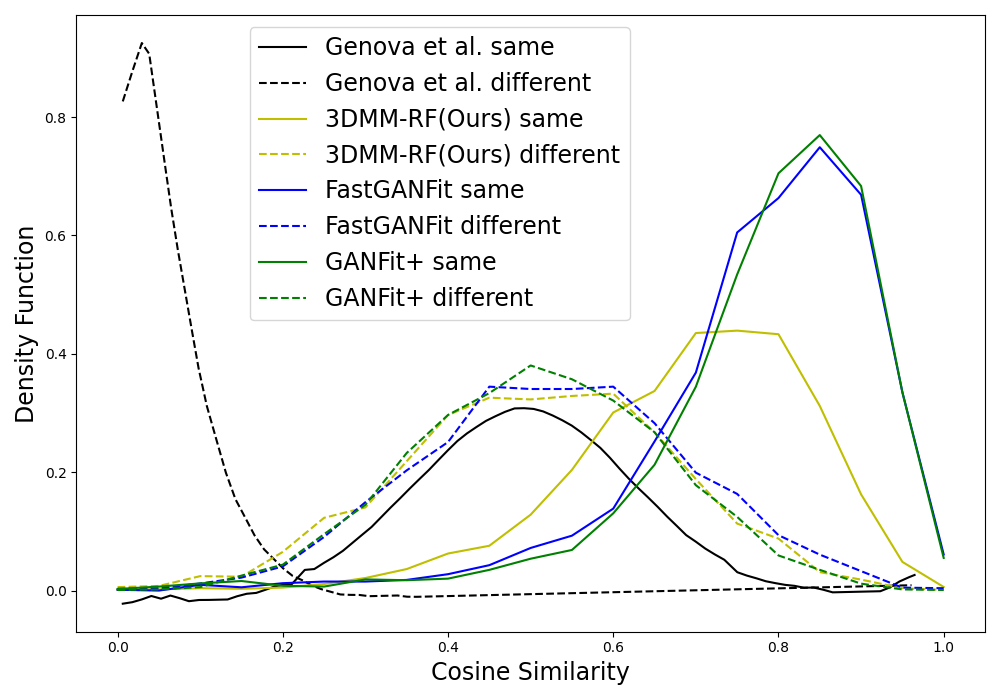}
         \caption{\centering Identity seperability of paired images}
         \label{fig:id_pairs}
     \end{subfigure}
\end{center}
\vspace{-0.35cm}
  \caption{\ref{fig:id_comparison}: The distribution of cosine similarity between the original images and the reconstructed-rendered images of LFW dataset, which indicates strong identity preservation of the reconstruction of our model. \ref{fig:id_pairs}: The distributions between same (solid line) and different (dashed line) pairs of LFW dataset after they have been reconstructed and rendered by our method. The separability of the two distributions shows that the identity is being preserved.}
  \vspace{-0.5cm}
\label{fig:ids}
\end{figure}

%------------------------------------------------------------------------
%  \vspace{-0.25cm}
\subsection{Ablation Study}

One of the key parts of our approach is the use of SPADE~\cite{park2019SPADE} layers for feeding the scene parameters to the network. We examine the importance of these, by implementing another approach without any SPADE layers, in which all the parameters are given together as input to the network. Fitting the image of Fig.~\ref{fig:ablation_gt} to both networks, we present our network's reconstructed image in Fig.~\ref{fig:all_loss} and a novel view of it in Fig.~\ref{fig:all_loss_side}, whilst Fig.~\ref{fig:wo_spade},\ref{fig:wo_spade_side} include the fitted results performed by the network without SPADE layers. Comparing those figures, it is easily shown that the \emph{w/o SPADE layers} network performs worse. We also notice that the resulting pose isn't the desired one, meaning that SPADE layers are a key factor to the disentanglement of the scene parameters.

Another ablation study focuses on the fitting pipeline. We examine the importance of each loss function and the finetuning final step by a leave-one-out experiment. As shown in Fig.~\ref{fig:ablation_study}, the results in Fig.~\ref{fig:all_loss} are the best in comparison with the others, meaning that the current used loss functions, the SPADE layers and the finetuning step play a key role to the performance of the approach.
Qualitatively, however, it's very difficult to distinguish the importance of the $L_{ID}$ loss. To measure it, we compare the cosine identity distance between the input image of Fig.~\ref{fig:ablation_gt} with the reconstructed images of Fig.~\ref{fig:all_loss} and Fig.~\ref{fig:no_id} using 
the identity detector provided by deepFace~\cite{serengil2020lightface}. We found that Fig.~\ref{fig:ablation_gt} is more similar to Fig.~\ref{fig:ablation_gt} than Fig~\ref{fig:no_id}, meaning that the $L_{ID}$ loss plays an important role in the fitting pipeline.
% , the former having identity distance equal with 0.104 and the latter 0.1337.
% This
\def \var {0.13}
\begin{figure}[t]
\begin{center}
  \centering
    \begin{subfigure}[t]{\var\textwidth}
         \centering
         \includegraphics[width=1.0\textwidth]{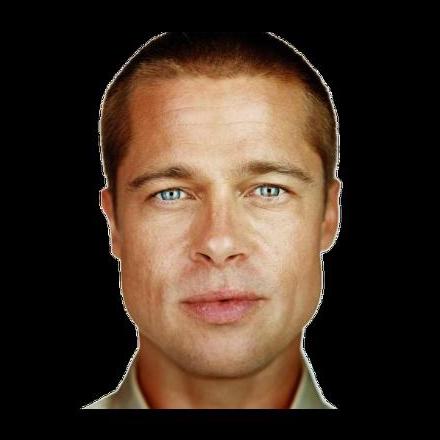}
         \caption{Input image}
         \label{fig:ablation_gt}
     \end{subfigure}
    \begin{subfigure}[t]{\var\textwidth}
     \centering
     \includegraphics[width=1.0\textwidth]{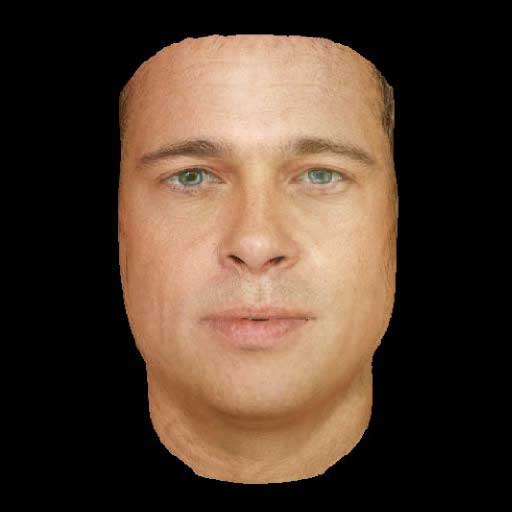}
         \caption{Ours}
         \label{fig:all_loss}
     \end{subfigure}
      \begin{subfigure}[t]{\var\textwidth}
     \centering
     \includegraphics[width=1.0\textwidth]{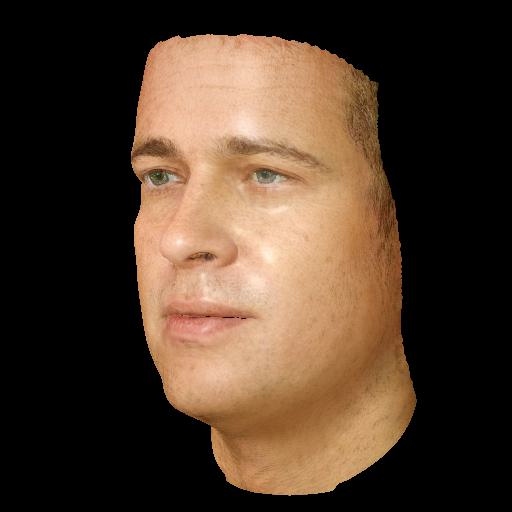}
         \caption{ \centering Ours ($\pi$/6)}
         \label{fig:all_loss_side}
     \end{subfigure}
     
     \begin{subfigure}[t]{\var\textwidth}
     \centering
     \includegraphics[width=1.0\textwidth]{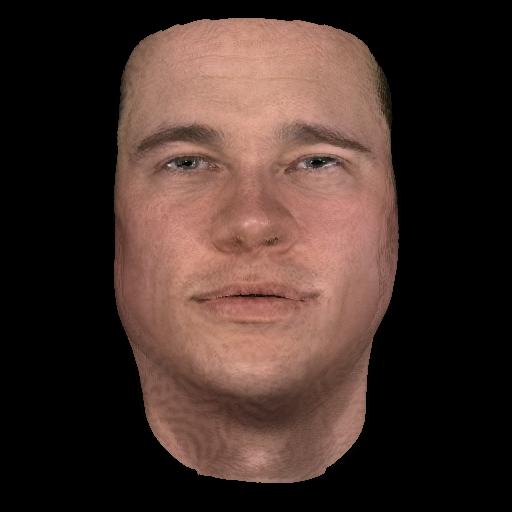}
         \caption{ \centering w/o SPADE}
         \label{fig:wo_spade}
     \end{subfigure}
     \begin{subfigure}[t]{\var\textwidth}
     \centering
     \includegraphics[width=1.0\textwidth]{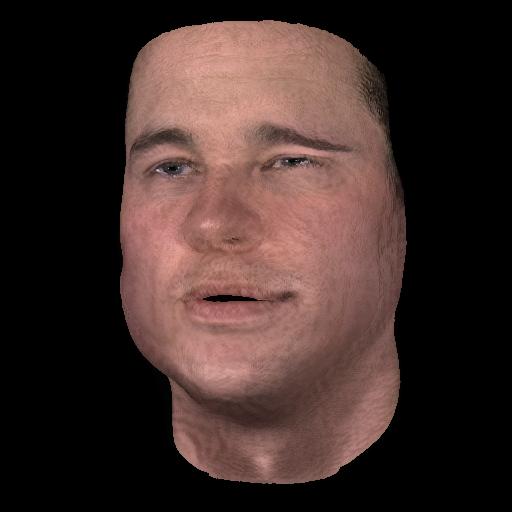}
         \caption{ \centering \footnotesize{} w/o SPADE ($\pi$/6)}
         \label{fig:wo_spade_side}
     \end{subfigure}    \begin{subfigure}[t]{\var\textwidth}
     \centering
     \includegraphics[width=1.0\textwidth]{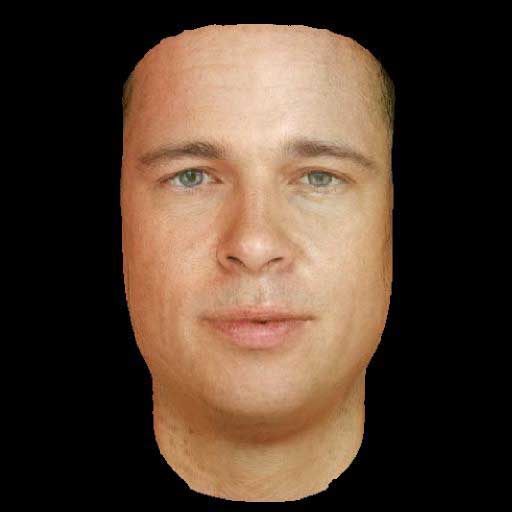}
         \caption{\centering w/o $L_{ID}$}
         \label{fig:no_id}
     \end{subfigure}
     
    \begin{subfigure}[t]{\var\textwidth}
     \centering
     \includegraphics[width=1.0\textwidth]{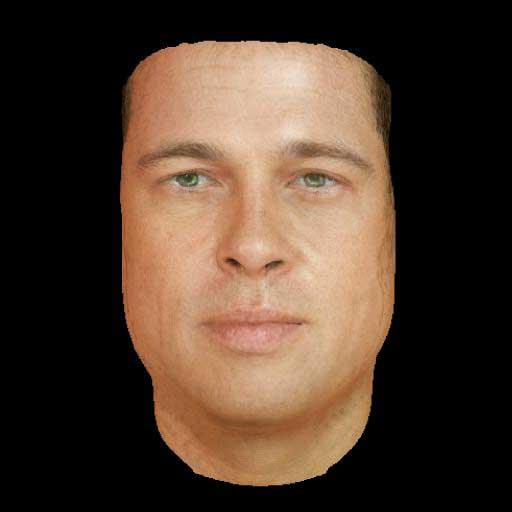}
         \caption{w/o $L_{pht}$}
         \label{fig:no_l2}
     \end{subfigure}
    \begin{subfigure}[t]{\var\textwidth}
     \centering
     \includegraphics[width=1.0\textwidth]{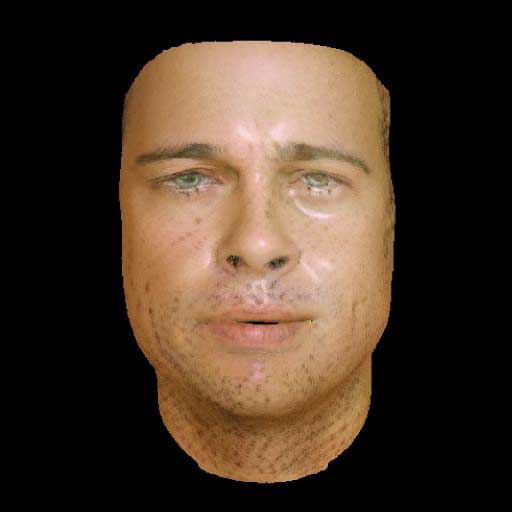}
         \caption{ \centering w/o $L_{vgg}$}
         \label{fig:no_vgg}
     \end{subfigure}
    \begin{subfigure}[t]{\var\textwidth}
     \centering
     \includegraphics[width=1.0\textwidth]{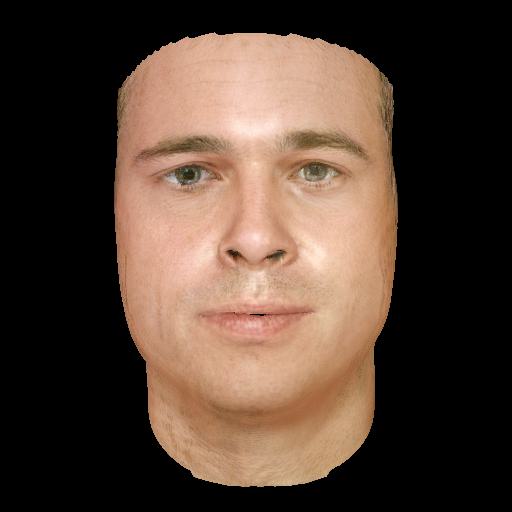}
         \caption{ \centering w/o Finetuning}
         \label{fig:no_mask}
     \end{subfigure}
     \vspace{-0.2cm}
\caption{Leave-one-out ablation study to visualize the contribution of various components of our fitting pipeline.
    % Fitting ablation study:
    % For (a) an input image,
    % we compare (b) the result of our method \ourname{},
    % with fitting variations:
    % (c) without identity loss $L_{ID}$,
    % (d) without photometric loss $L_{pht}$,
    % (e) without VGG loss $L_{vgg}$,
    % (f) without finetuning the network $\mathcal{G}$,
    % (g) without,
    % (h), and
    % (i).
}
\vspace{-0.5cm}
\label{fig:ablation_study}
\end{center}
\vspace{-0.5cm}
\end{figure}

%------------------------------------------------------------------------
\subsection{Limitations and Future Work}

Even though our network can achieve photo-realistic results, it still has some limitations. 
One of them is the fact that it cannot render other parts of the human head such as ears and hair. This is because the synthetic images we used for training don't include any of these parts. On top of that, in some situations where some hair occludes parts of the forehead, our network gets baffled. In future work, we think of using methods that are capable of producing human faces including all the parts.
Finally, the training data suffer from flattened eyes,
as they are represented by a single facial mesh,
which affects the final rendering.

Another limitation is the fact that our training images don't include any wearable like glasses. As a result, in some cases, our network didn't perform well in getting the right 
face texture because of the losses we are using during the fitting procedure. We consider finetuning our network in datasets including this type of data.

%------------------------------------------------------------------------
\section{Conclusion}
This work presents a facial deep 3D Morphable Model,
which can accurately model a subject's identity, pose and expression
and render it under arbitrary illumination.
This is achieved by utilizing a powerful deep style-based generator
to by pass two main weaknesses of neural radiance fields, rigidity and rendering speed,
by generating at one pass all the samples required for the volume rendering algorithm.
We have shown that this model can accurately be fit to ``in-the-wild'' facial images
of arbitrary pose and illumination conditions, extract the facial characteristics,
and be used to re-render the face under controllable conditions.

%%%%%%%%% REFERENCES
{\small
\bibliographystyle{ieee_fullname}
\bibliography{egbib}
}

\end{document}